%% file: csinet.tex
\documentclass{article}

\usepackage[preprint,nonatbib]{nips_2018}

\usepackage[utf8]{inputenc} % allow utf-8 input
\usepackage[T1]{fontenc}    % use 8-bit T1 fonts
\usepackage{hyperref}       % hyperlinks
\usepackage{url}            % simple URL typesetting
\usepackage{booktabs}       % professional-quality tables
\usepackage{amsfonts}       % blackboard math symbols
\usepackage{nicefrac}       % compact symbols for 1/2, etc.
\usepackage{microtype}      % microtypography

\usepackage{booktabs} % For formal tables
\usepackage{multirow}
\usepackage{caption}
\usepackage[normalem]{ulem}

\usepackage{tikz} %加水印用
\usepackage{eso-pic}
\usepackage{minitoc}

\usepackage{hyperref}

\usepackage[ruled]{algorithm2e} % For algorithms

\newcommand\ie{\textit{i.e.}\xspace}
\newcommand\etc{\textit{etc}\xspace}
\newcommand\eg{\textit{e.g.}\xspace}

\newcommand{\Figure}[1]{Fig.~\ref{fig:#1}}

\newcommand{\Sec}[1]{Section~\ref{sec:#1}}
\newcommand{\Table}[1]{Table~\ref{tab:#1}}

\newcommand{\net}[1]{CSI-Net}

\title{CSI-Net: Unified Human Body Characterization \\and Pose Recognition}

\author{
  Fei Wang\thanks{Work done when at Robotics Institute, CMU.} \\
  Xi'an Jiaotong University \\
  Carnegie Mellon University\\
  \texttt{feiwang@cmu.edu} \\
   \And
   Jinsong Han \\
  Xi'an Jiaotong University\\
  Zhejiang University \\
   \texttt{hanjinsong@zju.edu.cn} \\
   \And
   Shiyuan Zhang \\
   Xi'an Jiaotong University \\
   \texttt{shiyangzhang932@gmail.com} \\
   \And
   Xu He \\
   Xi'an Jiaotong University \\
   \texttt{hexu.xjtu@gmail.com} \\
    \And
   Dong Huang \\
   Carnegie Mellon University \\
   \texttt{donghuang@cmu.edu} \\
}

\begin{document}

\maketitle

\begin{abstract}
%   Channel State Information (CSI) of WiFi signals becomes increasingly attractive in human sensing applications due to the pervasiveness of WiFi, robustness to illumination and view points, and little privacy concern comparing to cameras.
%   In majority of existing works, CSI sequences are analyzed by traditional signal processing approaches. 
%   These approaches rely on strictly imposed assumption on propagation paths, reflection %penetration 
% and attenuation of signal interacting with human bodies and indoor background. This makes existing approaches very difficult to model the delicate body characteristics and activities in the real applications. To address these issues,
We build CSI-Net, a unified Deep Neural Network~(DNN), to learn representation of WiFi signals.
% , that fully utilizes the strength of deep feature representation and the power of existing DNN architectures for CSI-based human sensing problems. 
Using CSI-Net, we jointly solved two body characterization problems: biometrics estimation  (including body fat, muscle, water and bone rates) and person recognition. We also demonstrated the application of CSI-Net on two distinctive pose recognition tasks: the hand sign recognition (fine-scaled action of the hand) and falling detection (coarse-scaled motion of the body). 
% Besides the technical contribution of CSI-Net, we present major discoveries and insights on how the multi-frequency CSI signals are encoded and processed in DNNs.
Code has been made publicly at \url{https://github.com/geekfeiw/CSI-Net}.
\end{abstract}

\input{tex/introduction}

\input{tex/related_work}

\input{tex/csi_net}

\input{tex/implement_detail}
% \input{tex/net_tables}
\input{tex/results}

\section{Conclusion}\label{sec:conclusion}
In this paper, we propose a wireless signal model to analysis CSI variance at the presence of human body. We conclude that human presence even without doing anything can introduce CSI variance, which can has significant potentials for sensing human problems. We elaborately design a unified CNN based networks, named CSI-Net, to handle the CSI data for a variety of sensing human takes. We deploy our approaches to test the conclusion and CSI-Net on body characterization and activity recognition, including four tasks: biometrics estimation, person recognition, hand sign recognition and falling detection. The experimental results show that all tasks can be achieved with distinguished performance. Deserved to be mentioned, our work is the first attempt to propose and achieve biometrics estimation using commodity WiFi devices, which has applicable potentiality on health care. 
% Our work is also one of the earliest attempt to provide empirical understanding of the DNN behaviors in processing CSI data. 

\section*{Acknowledge}
We thank Wei Xi and Kun Zhao for discussion on WiFi properties. We thank Li Zhu, Pan Feng, Zhen Liao, Ziyi Dai and Yang Zi for their helps in data collection. Fei Wang is supported by China Scholarship Council.

{\small
\bibliographystyle{IEEEtran}
\bibliography{bibo}
}

\input{tex/appendix}

\end{document}

%% file: tex/introduction.tex
\section{Introduction}\label{sec:introduction}

Recent years witness rapid growth of techniques using Channel State Information (CSI) of WiFi signals in sensing human bodies. These techniques lead to major boost of applications in human-computer interaction \cite{qian2017inferring,li2016wifinger,ali2015keystroke}, health care\cite{fang2016bodyscan,wang2017rt,wang2017tensorbeat,wang2016human} and surveillance\cite{xi2014electronic,zheng2016smokey,adib2013see}. Comparing to computer vision based human sensing, WiFi-based techniques enable non-line-of-sight activity recognition\cite{adib2013see}, allow prevalent deployment in daily living and working environment, and arises little privacy concern.

The core problem of WiFi sensing is to prepare a proper representation of signals that directly correlates with human body characteristics and activities. 
Existing work used hand-crafted features:~(1) Temporally aligned CSI sequences by Dynamic Time Wrapping (DTW)\cite{palipana2018falldefi,abdelnasser2015wigest,abdelnasser2015ubibreathe, tan2016wifinger, ali2015keystroke,li2016csi}. (2) Statistical features of the CSI sequences, such as the average, variance, and entropy. It is unclear how to manually design person-specific features for body characteristics and person-invariant features for activities. Moreover, the interaction of WiFi signals with human body and background results in complex multi-path transmission, reflection and attenuation of CSI, making it extremely difficult for hand-craft feature to extract delicate information from CSI data.

On the other hand, recent advances of deep learning exhibit extraordinary ability in learning data representation from data. There are some preliminary attempts on indoor localization\cite{wang2017csi} and person recognition\cite{shi2017smart}. However, to our knowledge, no existing work has build a unified deep learning framework that solves multiple WiFi sensing problems.

The technical contribution of this paper can be summarized as follows.

% \begin{enumerate}
% \item 
1. We propose CSI-Net, a deep learning framework for human sensing with WiFi CSI sequences.

% \item 
2. We solve a variety of human sensing tasks using CSI-Net: biometrics estimation, person recognition, hand sign recognition and falling detection.

% \item 
3. We qualitatively analyze the influence of person body on WiFi signals and build a database for CSI-based human sensing tasks. 
% Based on extensive experiment results, we conclude that besides the tasks demonstrated in the paper, many more tasks, such as retrieving the human shape and pose,% \red{even viscus}, 
%  can be achieved using CSI-Net and WiFi signals. 
% \end{enumerate}

%%% sketch 1
\begin{figure}[t]
\includegraphics[width=0.8\linewidth]{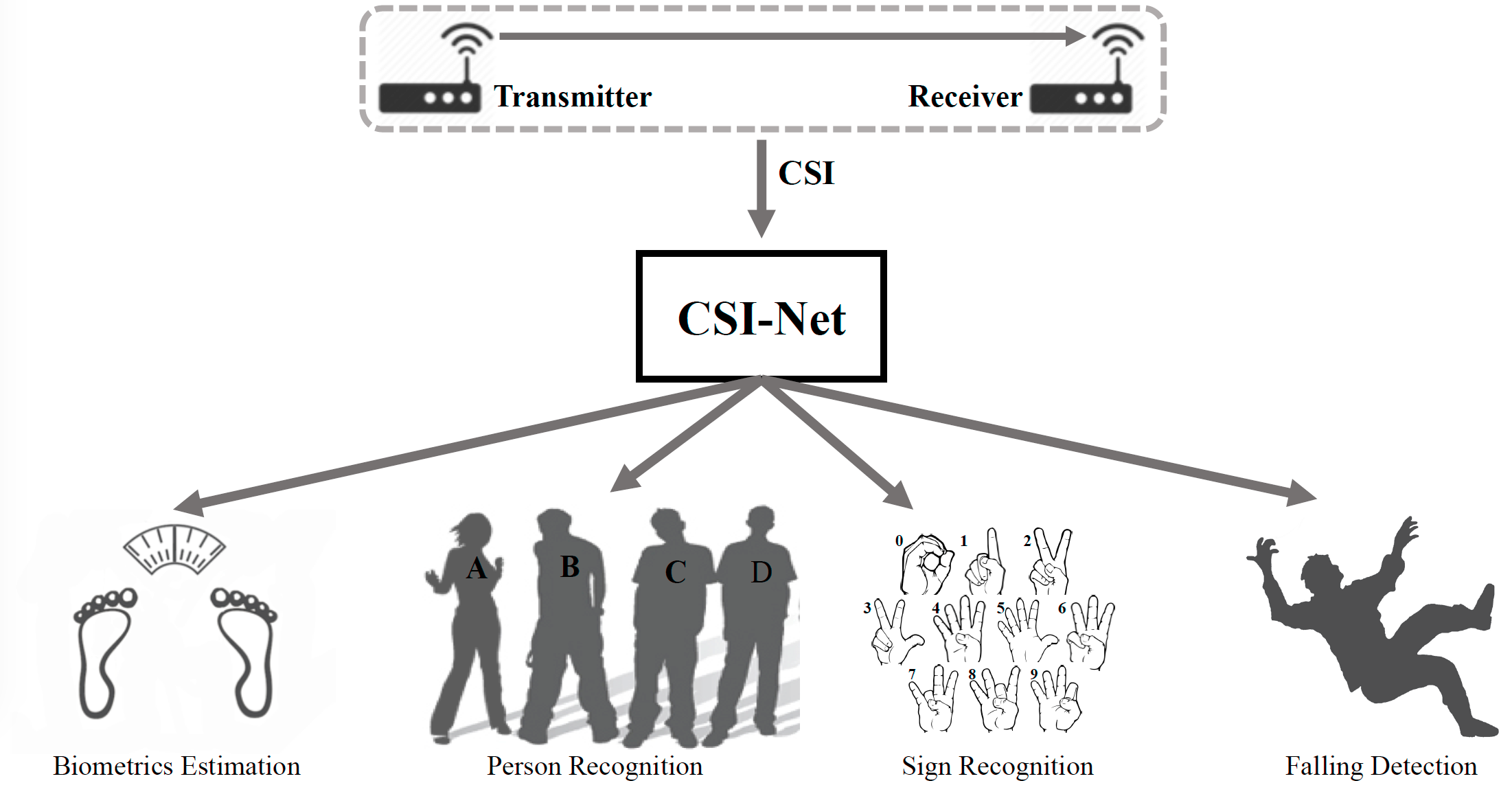}
   \centering
\caption{System overview. CSI is generated from the transmitter/receiver WiFi communication. CSI-Net is a unified body characterization and activity recognition deep learning architecture. In this paper, we apply CSI-Net to do body characterization tasks such as biometrics estimation and person recognition, meanwhile, we also apply it to do activity recognition tasks such as hand sign recognition and falling detection.
}
\label{fig:identification}
% \hspace{-20pt}
\end{figure}

%% file: tex/related_work.tex
\section{Related Work}\label{sec:related}
% Related work are roughly categorized according
% Related CSI-based work can be roughly categorized according to the method as well as the task.
% We summarize related CSI-based work on body characterization and activity recognition in this section.

\subsection{Body Characterization}\label{sec:prior_body}

The majority of CSI-based body characterization work focuses on person recognition \cite{wang2016gait,zeng2016wiwho,zhang2016wifi,xin2016freesense,lv2017wii,xu2017radio}. These work use walking pattern to identify person identity. Specifically, subjects are asked to walk along a predefined path repeatedly while the CSI sequences are recorded as training data. Then classification algorithms are developed to map the CSI data to identities. In these approaches, computing discriminative representation
for each user is very difficult yet essential to the identification performance. In \cite{zeng2016wiwho,zhang2016wifi,xin2016freesense,lv2017wii,xu2017radio}, statistical features, e.g., \textit{maximum}, \textit{minimum}, \textit{mean}, \textit{energy} and \textit{entropy}, are extracted from the CSI sequences and fed to classifiers, e.g., Support Vector Machine(SVM), for person recognition. In these work, feature representations of CSI are manually designed, and it is difficult to guarantee that features are discriminative for person recognition.

\subsection{Activity Recognition}\label{sec:prior_activity}

% There exist a wide spectrum of applications on activity recognition using WiFi signals. Besides walking, CSI sequences have been used to recognize keystroking\cite{ali2015keystroke,li2016csi}, breathing\cite{wang2016human,liu2016contactless} and falling\cite{wang2017wifall,wang2017rt,palipana2018falldefi}. 
In terms of processing flow, recent approaches in CSI-based activity recognition fall into two main schemas. \textbf{(1) Feature classification}: \textit{raw CSI} -> \textit{filtered CSI} -> \textit{hand-crafted features} (statistical features)-> \textit{classifier} -> \textit{activity classes}~\cite{wang2017wifall,wang2017rt}.
% This schema uses statistical features, such as \textit{maximum, minimum, mean, energy and entropy}, to represent variance of CSI sequences caused by body activities %\cite{wang2017wifall,wang2017rt,palipana2018falldefi}
% and train a classifier to recognize these activities. However, some researchers showed that an activity may vary in duration, speed or range even for a same subject\cite{ali2015keystroke}, which deceases the robustness of recognition. 
\textbf{(2) Sequence matching}: \textit{raw CSI} -> \textit{filtered CSI} -> \textit{dynamic time wrapping (DTW)}~\cite{berndt1994using} -> \textit{k-nearest neighbors (kNN)} -> \textit{activity prediction}. 
% This schema directly computes the distance of CSI variance aligned by DWT and employs kNN to recognize the activities with minimal distance\cite{tan2016wifinger, ali2015keystroke,li2016csi,palipana2018falldefi,abdelnasser2015wigest,abdelnasser2015ubibreathe}. 
% The second schema may perform better in discriminating activities consisting of micro activities like keystroke\cite{ali2015keystroke}. 
% Similar to the hand-craft features, the performance of kNN rely solely on manually designed distance measures of CSI sequences.
% However, to recognize an activity, 
The former schema has shortages in designing task-specifically features.
The later schema requires computing distances between a testing CSI and every training CSI, which is very time-consuming. 
In addition, kNN algorithm is generally inferior to a supervised learning classifier.

\subsection{Bio-electromagnetics in WiFi Signal Band}\label{sec:bio} 

Our work is based on fundamental principles in Bio-electromagnetics.  Bio-electromagnetics studies the  changes of electromagnetic~(EM) waves when encountering biological bodies. Many literatures specifically studied the Bio-electromagnetics in WiFi Signal Band, \eg, 2.4G WiFi (2470MHz-2544MHz) and 5G WiFi~(5033MHz-6006MHz). In \cite{gabriel1996dielectric}, dielectric parameters like permittivity and conductivity of body tissues, such as fat, muscle and liver \etc, were measured within the EM frequency of 10Hz to 20GHz. 
% (see details in \Table{tissue}).
In \cite{christ2006characterization, christ2006dependence, wang2018wipin}, human body is modeled as layered tissues with different dielectric parameters.
% , as sketched in \Figure{body}~(right). 
% Specific absorption rate~(SAR) of WiFi signal propagating through these layered tissues can be measured and used to characterize human body biometrics. Furthermore, standing-wave effects occur at certain fat layer thickness and frequencies within 30MHz-6000MHz.
\cite{dove2014analysis} also used the layered tissue model to compute the power decay and time delay of EM wave when propagating through human body at the frequency of 2.45GHz. This work provides critical cues to measure human body spatially, making it possible to capture body gestures/actions using the decayed and delayed WiFi signals.

%% file: tex/csi_net.tex
\section{Network Architecture}\label{sec:net}

\begin{figure*}[t]
\includegraphics[width=0.8\linewidth]{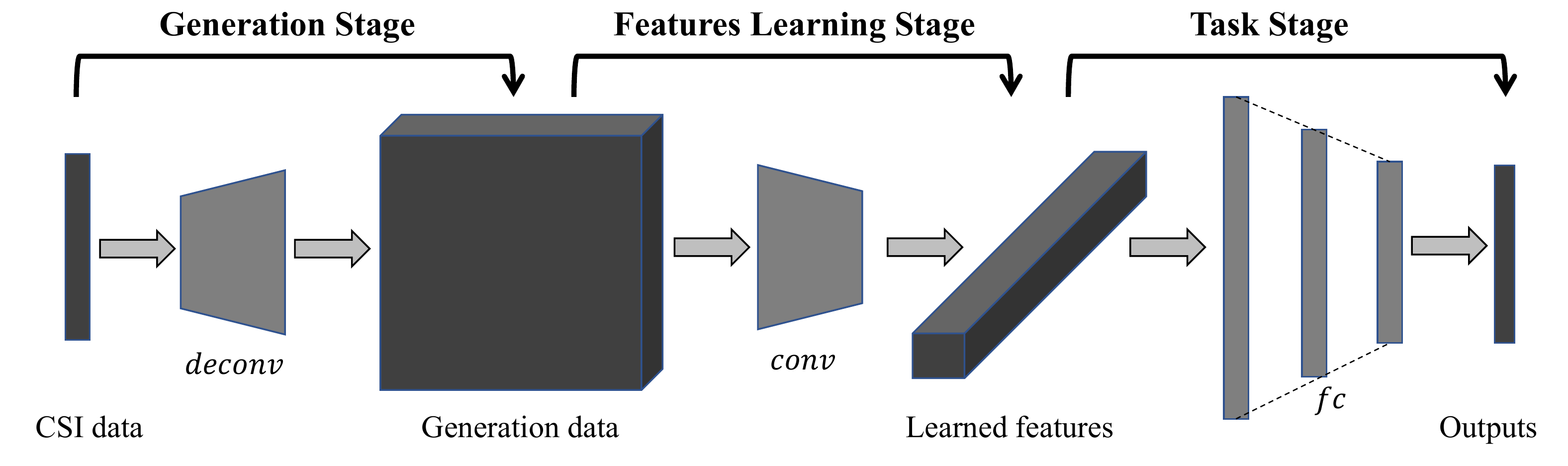}
   \centering
\caption{Architecture of CSI-Net. 
% ``Generation Stage'' transforms CSI sequence to spatially-encoded maps. ``Features Learning Stage'' extracts features from spatially-encoded maps. ``Task Stage'' map features to the output of individual tasks. CSI-Net can handle multiple tasks simultaneously by parallelizing multiple task stages after the shared ``Generation Stage'' and ``Features Learning Stage''.
}
\label{fig:csi_net}
% \hspace{-20pt}
\end{figure*}

% The architecture of CSI-Net is showed showed at \Figure{csi_net}, 
As shown in \Figure{csi_net}, CSI-Net is composed of three functional components: the generation stage, the feature learning stage and the task stage. The generation stage transforms CSI data to feature maps with the spatially-encoded patterns. The feature learning stage further maps the spatially-encoded patterns to the features for sensing tasks. Finally, the features are used to produce outputs in the task stage.
  
% In the remainder of this section, we first describe the input tensor of CSI-Net, then explain three functional components in detail.

\subsection{Input Tensor of CSI-Net}\label{sec:input}
CSI is computed from a channel estimation process in the 802.11 n/g WiFi system. 
% Given a WiFi packet from a transmitter to a receiver,
% channel estimation is formulated as,
% \begin{equation}
% % Y_k = H_kX_k + N_k, k \in [1, S_c]
% Y = HX + n
% \end{equation}
% where, $Y$ is the received data, $X$ is the transmitted data, $H$ is the CSI, and $n$ is the noise of the WiFi system. 
Using \cite{halperin2011tool}, we can extract CSI as a complex-valued sequence, $\mathbf{A}\in \mathcal{C}^{N_{sa}\times N_{sc}\times N_{tx} \times N_{rx}}$, where $N_{sa}$ and $N_{sc}$ are the number of sampled WiFi packets and number of subcarriers, respectively. $N_{tx}$ and $N_{rx}$ are number of transmitting and receiving antennas, respectively. In our setting, $N_{sc}=30$, $N_{tx} = 1$ and $N_{rx}=1$. 
 % In our experiments, we deploy Linux 802.11n CSI Tool~\cite{halperin2011tool} to record CSI sequences of $30$ subcarriers. 
% Due to the unstable CSI phases, we only use CSI amplitudes. 
% as raw input data $\mathbf{A}\in \mathcal{R}^{N_{sa}\times 30~\times 1 \times 1}$,~$\mathcal{R}$ denotes the set of \textit{real} numbers. %\marginpar{Review 3, define CSI}

% \begin{table}[th]
% \centering
% \begin{tabular}{lll}
% \hline
% Tool                      & Protocol& Subcarrier number       \\ \hline
% Linux 802.11n CSI Tool\cite{halperin2011tool}   & 802.11n  & 30 \\  
% USRP+GNU Radio\cite{yubo2013zimo}             & 802.11g  & 52(data)+4(pilots)                    \\
% Atheros CSI Tool\cite{xie2015precise} & 802.11n  & 56(20MHz), 114(40MHz) \\ \hline
% \end{tabular}
% \caption{Three software tools that produces CSI of multiple subcarriers. The number of subcarriers corresponds to the $channel$ number of the input tensor of CSI-Net.  }
% \label{tab:csi_tool}
% \end{table}

% \marginpar{Review 3, incomplete description of how CSI data is mapped to the input layer.}

% Note that, in conventional computer vision problems, input tensors of Deep Neural Networks (DNNs) are images with size of $Channel\times Height \times Width$, where $Channel$ is the number of color channels, $Height$ and $Width$ represent the image height and width, respectively. Typical RGB images have 3 channels (red, green and blue), which correspond to visual information of the same natural world at three different frequency sets. Similarity, WiFi signals sense the same natural world at $N_{sc}$ electromagnetic frequencies (\ie, $N_{sc}$ subcarriers). 
% Moreover, from the successful multi-channel signal models, such as blind signal separation (BSS)~\cite{amari1996new}, the combined signals can be well-approximated as a linear combination among channels. We keep the interaction among channels while release the information within individual channels using convolution computation.
We take the CSI at each sampling time-stamp as a multi-channel tensor, ($a_i\in \mathcal{C}^{30\times 1\times \times 1}, i\in[1,N_{sa}]$). Due to the unstable CSI phases, we only use CSI amplitudes.

\subsection{Generation Stage} \label{sec:generation}
% Recent DNN architectures in computer vision community are very powerful and efficient in processing spatial information encoded within image channels. However, there are rare case studies that directly apply DNNs to the $\mathcal{R}^{ 30 \times 1\times 1}$ input tensors. % Considering that the size of CSI-Net inputs is quite smaller than traditional CNN inputs, images, we aim to enlarge the size of inputs to take the advantages of modern and efficient CNN architectures in processing images. 

Using transposed convolution (TC), also called deconvolution, we build a generation stage of CSI-Net. The generation stage explore the within-channel for spatial information CSI-Net input tensors while make it possible to take advantages of modern DNN architectures on images. TC is widely used in Generative Adversarial Nets (GANs) to for image generation \cite{goodfellow2014generative, salimans2016improved}. 
Specifically, We here stack 8 transposed convolution layers 
% (7/8: kernel size of $4 \times 4$, 1/8: kernel size of $3 \times 3$) 
to process the $30 \times 1 \times 1$ input tensor, and produce a tensor of $6\times 224 \times 224$. The target tensor size is set to $224\times 224$ is because it is a common input size of CNN architectures
excluding VGG-nets
\cite{simonyan2014very} , Inception\cite{szegedy2015going} and ResNets\cite{he2016deep}, hence easier in tuning hyperparameters. 

\subsection{Features Learning Stage}\label{sec:feature_learning}
Feature learning stage takes the $6\times 224 \times 224$ tensor from the generation stage and extracts features for final tasks. Following the fashion in computer vision, we call the networks in  feature learning stage as the backbone network. Given image-like tensors after the generation stage, various modern CNN architectures, such as AlexNet~\cite{krizhevsky2012imagenet}, VGG-nets\cite{simonyan2014very}, Inception clusters\cite{szegedy2015going, ioffe2015batch, szegedy2016rethinking, szegedy2017inception}, ResNets\cite{he2016deep}, and DenseNet~\cite{huang2017densely} can be potentially used as backbone with slight revisions. Following insights in \cite{canziani2016analysis}, we select ResNet for it is generally more powerful than AlexNet, has less parameters than VGG-nets, Inception Net and DenseNet. 

\subsection{Task Stage}
The task stage is to leverage features learned above to compute the outputs for a specific task. It is composed of several fully-connected layers, activation function and loss function. 

Fully-connected layers re-weight the features to produce an output vector for a task. As shown in \Table{task_output}, to estimate 4 biometrics, \ie, human body fat rate, muscle rate, water rate and bone rate, the output dimension is 4. Person recognition is conducted among 30 subjects, so the output dimension is 30. 
% Because the fully-connected layer has much more parameters than convolution layers, which increases the complexity of CSI-Net and difficulty in optimization, we tend to use more convolution layers than fully-connected layers. 
In our experiments, we only use one fully-connected layer.

% However, the parameter of fully-connected layer is $O(mn)$, where $m$ and $n$ is the 

% Activation function and loss function are used to compute 
% the loss in training CSI-Net, which measures the difference between the network outputs and the ground-truth values. 
In the biometrics estimation task, we use Rectified Linear Unit (ReLU)\cite{nair2010rectified} plus L1 to compute loss.
%  We select ReLU is because biometrics is always greater than zero and 
In other tasks, the SoftMax activation and Cross Entropy function are combined to generate loss.

% Multiple \textit{fully-connected} layers are stacked in this stage to solve specific tasks. What costs is to decide the amount of layers and the neuron amount of each layer. After that, \textit{activation function} and \textit{loss function} are applied on the outputs of the last layer to compute the \textbf{loss} of the deep networks for back-propagation optimization. 
% Table 2 lists two most common \textit{activation-loss pairs} for  classification and regression tasks. 

% \begin{table}[]
% \centering
% \caption{Common activation-loss pair for classification and regression}
% \label{my-label}
% \begin{tabular}{l|l|l}
% \hline
%  Task & Activation  & Loss                               \\ \hline
% Classification        & Sigmoid       & Log loss                      \\ \hline
% Regression           & ReLU          & L1-loss, L2-loss, MSE-loss... \\ \hline
% \end{tabular}
% \end{table}

% CSI-net is capable to cope with multiple tasks simultaneously with corresponding task stages like Faster R-CNN\cite{ren2015faster}. Thus, the \textbf{loss} of networks is the sum of \textbf{loss} of each tasks, which formulated as,
% \begin{equation}
% L = \alpha_1 l_1+\alpha_2 l_2+ ...+ \alpha_nl_n
% \end{equation}
% where $\alpha_i (i\in n)$ is the \textit{balance} among losses.

\begin{table}[h]
\centering
\scriptsize{
\begin{tabular}{l|l|l|l|l}
\hline
 Data Sizes / Tasks                   & Biometrics Estimation    & Person Recognition      & Sign Recognition          & Falling Detection         \\ \hline\hline
CSI-Net Input             & $30 \times1\times1 $       & $30 \times1\times1 $       & $30 \times1\times1 $       & $30 \times1\times1 $       \\ \hline \hline 
Generation Stage    & $6 \times 224 \times 224$ & $6 \times 224 \times 224$ & $6 \times 224 \times 224$ & $6 \times 224 \times 224$ \\ \hline
Feature Learning Stage& $256$                  & $256$                      & $128$                       & $128$                       \\ \hline
Task Stage             & $4$                         & $30$                        & $10$                        & $2$                         \\ \hline\hline
% Activation          & ReLU                      & SoftMax                   & SoftMax                      & SoftMax                      \\ \hline
% Loss       & L1                        & CrossEntropy              & CrossEntropy                        & CrossEntropy                        \\ \hline

CSI-Net Output       & $4$                         & $30$                        & $10$                        & $2$                         \\ \hline
\end{tabular}}
\vspace{5pt}
\caption{Tensor sizes at different stages of CSI-Net.
% Four sensing tasks have the same input size $channel \times height \times width  : 30 \times 1 \times 1$. After generation stage, the spatially-encoded maps are of size $6\times 224 \times224$. The outputs of feature learning stage were set differently for the best performance of each task. 
% The output sizes at the task stages are task-specific: Output size $4$ for four scalars in biometrics estimation; Output size $30$ for $30$ subjects used in person recognition; Output size $10$ for hand signs encoded by $0$ to $9$; Output size $2$ for falling is detected or not detected. The final output size of CSI-Net is the same as output size of task stages.
    % , meanwhile, biometrics prediction task utilizes ReLU and L1 to compute \textit{loss} while the other three use SofaMax plus CrossEntrop
}    
\label{tab:task_output}
\end{table}

\subsection{Configuration on Different Tasks}
For different sensing tasks, the output functions and intermediate tensors are different, as
illustrated in \Table{task_output}. Taking the task of biometrics estimation as an example, the input tensor size is $Channel\times Height \times Width  : 30 \times 1 \times 1$. After the generation stage, the tensor is $6 \times 224\times 224$. Then,
$256$ features are extracted by the feature learning stage. At last, the task stage estimate four biometrics (4 scalars) from the $256$ features.
The output tensor size of CSI-Net is equal to the output size of task stage.
More implementation details of CSI-Net on different sensing tasks are presented in \Sec{detail}.

%% file: tex/implement_detail.tex
\section{Implementation Details}\label{sec:detail}

In this section, we present the implementation details of CSI-Net, including the preparing dataset and training CSI-Net.

% As section.model said, the human body affects WiFi signals and it is intuitive that this effect varies among persons for the uniqueness of their body shapes and internal structures. Based on this intuition, a direct proposal is to identify a person. Another proposal is to estimate biological information of a person, here we estimate body fat, muscle, water and bone rate because these values are easier to collected through a electronic fat scale. 

% We recruit 30 volunteers whose information is listed at Table 4 for the proposals. Every one is asked to stand between the transmitter and receiver for about 1-2 minutes with CSI recored concurrently. For focusing on algorithm, we leave the details of experimental setup and data collection at the section.data.

% \begin{figure*}[h]
% \includegraphics[width=0.7\linewidth]{figs/layout.pdf}
%   \centering
% \caption{Data collection site snapshot and room layout. Figures from top to down are for biometrics estimation/person recognition, hand sign recognition and falling detection. In biometrics estimation, a pasted cardboard and a brick are used to limit the position and orientation of subjects. Setting in sign recognition is an imitation of \cite{li2016wifinger}. In falling detection, we collected data from 5 selected positions, shown in gray at the lower right layout figure.}
% \label{fig:layout}
% % \hspace{-20pt}
% \end{figure*}

\begin{table}[ht]
\centering
\scriptsize{
\begin{tabular}{l|l|l|l|l}
\hline
 System Settings/ Tasks                & Biometrics Estimation & Person Recognition & Sign Recognition & Falling Detection \\ \hline \hline
Tx-Rx Height            & 1.2m                   & 1.2m                 & 0.8m             & 0.8m              \\ \hline
Tx-Rx Distance          & 1.6m                   & 1.6m                 & 0.6m             & 3.0m              \\ \hline
\# of Subjects       & 30                     & 30                   & 1                & 1                 \\ \hline
Subject Body Pose       & standing                     & standing                   & siting                & standing\&laying                 \\ \hline

\# of Positions        & 1                      & 1                    & 1                & 5                 \\ \hline
Sampling Duration & 100s                   & 100s                 & 60s              & 30s               \\ \hline
Sampling Rate     & 100Hz                  & 100Hz                & 100Hz            & 100Hz             \\ \hline
% \#Antenna         & 1tx-1rx                & 1tx-1rx              & 1tx-1rx          & 1tx-1rx           \\ \hline
CSI Samples         & 300K                & 300K              & 60K          & 30K          \\ \hline
Center Frequency
         & 5GHz                & 5GHz              & 5GHz          & 5GHz          \\ \hline

\end{tabular}}
\vspace{5pt}
\caption{System settings in experiments. 
% Taking biometrics estimation for example, the transmitter (``Tx'') and receiver (``Rx'') are put on two boxes with height of $1.2m$ from the floor (``Tx-Rx Height'') and distance (``Tx-Rx Distance'') of $1.6m$ . We asked 30 subjects to be ``standing''  in the room. The CSI data were recorded for 100 seconds with sampling rate of 100Hz. The total number of CSI samples is 300k ($30\times100\times100$). The transmitter and receiver were set to work at 5GHz WiFi band.
}
\label{tab:deployment}
\end{table}

% In this section, we describe the implementation details of each task on 
% \begin{itemize}
% \item The way to prepare dataset;
% \item The way to train the networks.
% \end{itemize}

\subsection{Preparing Datasets}
The dataset preparation consists of testbed setup, CSI collection, filter design, training/test data split and data augmentation.  

\subsubsection{Testbed setup}
%deploy?
We employed Commodity Off-The-Shelf (COTS) devices to setup our testbed. 
We replaced the network interface card (NIC) with Intel 5300 NIC at 2 mini-PCs. 
% (32G SSD, 1G memory, Ubuntu 14.04 OS). 
% One mini-PC worked as the WiFi signal transmitter~(Tx), while the other worked as the receiver~(Rx).
Both of them were installed with Linux 802.11n CSI Tool~\cite{halperin2011tool} for parsing CSI. 
% The Tx and Rx were deployed in a room (about $5m\times 6m$) with ordinary furnitures such as tables, chairs and desks. 
More setup details are listed in \Table{deployment}.

% we deploy them in a room($5m\times6m$) with different heights and distances for our tasks under considerations as follows.

% \begin{itemize}
% \item Height: the height of abdomen or chest approximates $1.2m$ for most people, as the model showed in \Figure{body}, we speculate that abdomen and chest would convey most abundant human information in CSI, so we use $1.2m$ in the task of biometrics prediction and human identification. For volunteer asked to sit at hand sign language recognition, his hand is about with height of $0.8m$ when being lifted. In the falling detection task, CSI should reflect the state when he standing as well as laying on the ground, so the height of $0.8m$ is a compromised height.

% \item Distance: we set the distance between Tx-Rx $1.6m$ is 
% Because hand number sign is a macro activity, which 
% \end{itemize}

\subsubsection{CSI data collection}\label{sec:collection}

% We collected data with IRB approval. \red{The majority of related prior work require subjects to repeat each activity tens of times~\cite{ali2015keystroke,wang2017wifall,wang2017rt,tan2016wifinger,zeng2016wiwho,xin2016freesense} for preparing datasets, but it is counterintuitive to train CSI-Net with small datasets. Besides, it is very inconvenient to ask subjects to repeat each activity hundreds of thousands of times to prepare a large dataset. Thus, we adopt a compromised data collecting scheme. That is,} 

CSI samples were recorded when subjects keep stationary poses or gestures during data collection. The CSI sampling rate was set to 100Hz for all tasks below.

% \begin{enumerate}

% \item  
\textbf{Biometrics Estimation:} we recruited 30 subjects, measured some biometrics of their bodies with Mi\textregistered ~
 body composition scale\cite{mibodysscale}, and listed the information in the \Table{person}. We selectively estimated the body fat rate and body muscle rate because these two metrics indicates what the body shape a person has. We also estimated the water rate and bone rate, which are essentially related to the human body internal composition. 
 
% As shown in \Figure{layout}, subjects were asked to stand at a predefined position. To avoid potential CSI biases caused by human position and orientation, we marked the standing position with a brick putting on the pasted cardboard, as shown in \Figure{layout}. To avoid impacts from the presence of other persons in the environment, we first turned on the CSI tool in the Tx and Rx to record CSI series, while keeping the room without any person. Then a subject walked to the marked position and kept stationary pose. 100-second CSI series, or about 10k samples (at 100Hz), were collected for each subject. We performed this procedure for 30 subjects and generated around 300k CSI samples for the biometrics estimation task. During the dataset collection, we did not change controllable environment settings, such as the height of Tx and Rx, distance between Tx and Rx, placement of furnitures, \etc.
 
% \item 
\textbf{Person Recognition:} we 
% use the same data as 
reused the sampled CSI for biometrics estimation to identify human. Actually, we trained a CSI-Net with 2 separate task stages, one for biometrics estimation and another for person recognition, with the same inputs.  
% Note that in the experiment, each subject was standing still. Thus, the identity is solely based on their unique biometrics.

% \item 
\textbf{Sign Recognition:}
for a proper comparison with non-adapted DNNs work, we imitated the experiment setting shown in a prior work~\cite{li2016wifinger} to collect CSI dataset for hand sign recognition. The imitations include the same selection of hand signs, height of Tx and Rx, and distance between Tx and Rx, even similarly asking subjects to sit on a chair \etc. 
% In particular, in our experiment one subject was asked to sit on a chair to pose 10 signs sketched at \Figure{finger}  (one for 60 seconds). 
%Each sign represents a number. We reduced the height of Tx-Rx($0.8m$) to fit the height of the subject's hand when sitting. Meanwhile, we decreased the distance~($0.6m$) between Tx and Rx to enhance the sensitivity of CSI variance on different postures, such that micro and similar hand signs can be distinguished. We also kept the controllable surrounding environment unchanged with the same requirement depicted in the biometrics estimation part.

% for the consideration of 
% , for is because that these pose is very similar,

% \begin{figure*}[t]
% \includegraphics[width=0.9\linewidth]{figs/finger.pdf}
%   \centering
% \caption{American sign language from 0 to 9 \cite{fingerasl}.}
% \label{fig:finger}
% % \hspace{-20pt}
% \end{figure*}

% \item 
\textbf{Falling Detection:} we selected 5 positions in the room and asked one subject to act action of standing and laying (the duration of each action is about 30 seconds), where the laying was to simulate the human body state of fallen down. 

\subsubsection{Training/Testing sets.}  \label{sec:dataset}  

% Note that CSI sequences drift along time \Figure{filter} due to several factors: First, the Intel 5300 NIC is imperfect and severely suffers from noises. The impacts include uncertainty in the packet boundary detection and mismatch in sampling frequency \cite{xie2015precise}. Second, human heartbeat and respiration may introduce periodic noises~\cite{wang2017tensorbeat,wang2016human}. The last but not the least, it is very hard for subjects to keep completely stationary for more than 100 seconds~(CSI collecting duration of biometric estimation and person recognition task), although they have been asked to keep stationary. Randomly splitting training and testing is equivalent to use all data for training and testing.

% \marginpar{Review 4, In \Figure{filter}, why signal amplitude varies over time when person is stationary.}

% As illustrated in the last subfigure of \Figure{filter}, 
In all tasks, we used the first 4/5 CSI sequences for training and the remaining 1/5 for testing. We took this splitting strategy to simulate the practical scenarios, where one first collects data and trains a network, then tests incoming data.
% \marginpar{Review 4 and Review 5, how training samples were selected?}

% To make CSI-Net more robust to CSI noise, before to train CSI-Net, 
% we further perform two operations on CSI sequences:
% \begin{itemize}
% \item Local Temporal Average: To smooth temporal random noise, we compute the average of every 10 continuous CSI samples with stride of 10, and use the average value as one input sample for CSI-Net.
% \item
% Augmenting Training Data: 
% Due to the drift of CSI sequences. CSI-Net would perform well in training data but fail on test data, which is known as the overfitting problem. 
We propose a data augmentation method, shown in Algorithm~\ref{alg:augmentation}. 
The main idea is to deliberately introduce some jitter CSI to the training data, so that CSI-Net can learn to model more data variances.
% \marginpar{Review 4, provide more description in text to explain Algorithm 1. }
% Specifically, denote the original training set as $S$, in which samples are from $s_1$ to $s_n$. The  data augmentation algorithm runs in multiple iterations. In a certain iteration, for example the iteration of $k=2$, the instances in $S$ are randomly shuffled. Then, we compute the averages of every 2 adjacent instances with stride of 2, and input them as new instances into augmented dataset $S'$. We iteratively perform this operation 4 times~($k=2,3,5,7$). At last, combing the augmented items with the original training set generates the augmentation set to train networks. 
% The key operation of this algorithm is averaging the shuffled training instances, which makes CSI-Net to be trained with much more CSI patterns.
Thus, it would be more resilient to the unpredictable CSI in the test phase. 
% In our experiments, this augmentation approach significantly increased the person recognition accuracy from $78.26\%$ to $93.00\%$. The numbers of training and test samples are listed in \Table{datasize}.

% \end{itemize}
% \marginpar{Review 4 and Review 5, how many samples were selected for training and testing.}

% We applied above two operations for the dataset of all tasks. The numbers of training and test samples are listed in \Table{datasize}.
% for biometrics estimation, person recognition, hand sign recognition and falling detection are 38,633, 38,633, 21,428 and 21,879, respectively. These numbers are sufficiently big for training our deep neural network, \ie, CSI-Net. The numbers of test instances for 4 tasks are also listed in \Table{datasize}.
% \marginpar{Review 5, amount of training and testing data}

\begin{algorithm}[h]
\caption{\bf Data Augmentation} 
\label{alg:augmentation}
\KwIn{Original Training Set: $S = \left \{ s_1, s_2, ..., s_n \right \} $}
\KwOut{Augmented Training Set: ${S}'$}
Initialize $S'=\left \{ \right \}$

\For {$k \in[2,3,5,7]$}
{Shuffle $S \rightarrow {S}^* = \left \{ {s}^*_1, {s}^*_2, ..., {s}^*_n \right \} $ \\
Initialize $S_k=\left \{ \right \}$\\

\For {$i \in [1,2,...,\left \lfloor n/k \right \rfloor]$}
{
Average $k$ continuous shuffled instances $\frac{1}{k}\sum_{(i-1)k+1}^{ik}{s}^*_i \rightarrow {s}'_i$\\

$    {s}'_i \cup S_k   \rightarrow S_k$\\
}

$ {S}_k \cup S'  \rightarrow S'$
}  
$S' \cup S \rightarrow {S}'$
\end{algorithm}

\begin{table}[h]
\centering
\scriptsize{
\begin{tabular}{l|l|l|l|l}
\hline
      Sets /Tasks            & Biometrics Estimation & Person Recognition & Sign Recognition & Falling Detection \\ \hline\hline
\#Training instances & 38,633                      & 38,633                    & 21,428                & 21,879                 \\ \hline
\#Test instances    & 4,444                   & 4,444                   & 2,468                & 2,519                 \\ \hline
\end{tabular}}
\vspace{5pt}
\caption{The training/test ``Sets'' for four sensing tasks. For example, the tasks of biometrics estimation and person recognition used $38,633$ training instances and $4,444$ test instances, respectively.}
\label{tab:datasize}
\end{table}

\subsection{Loss Function}

% \begin{itemize}
% \item
Biometrics Estimation and Person Recognition: We set the $loss$ of the jointly-trained tasks, denoted with $L$, as follows,

\begin{equation}
L = L_{bio} +\alpha L_{id}
\end{equation}
where $L_{bio}$ and $L_{id}$ are the \textit{losses} of biometrics estimation and person recognition, respectively. $\alpha$ works as a importance balance between both tasks. $L_{bio}$ is the sum of 4 biometrics estimation $losses$ that are computed by L1 Loss, and the $L_{id}$ is computed by the Cross Entropy Loss. Following settings in Fast R-CNN\cite{girshick2015fast}, we normalize the biometrics values within [0,1] before training, meanwhile, we set the balance $\alpha$ to 1.
% , which means we equally treat the importance of both tasks.

% \item
Sign Recognition: We compute the $loss$ of hand sign recognition by 
\begin{equation}
L = -\log p_i
\end{equation}
where $p_i$ is the $i$-th SoftMax result of an instance that is with a true label of $i$. Actually, this is one of writing styles of the Cross Entropy Loss.

% \item 
Falling Detection: The Cross Entropy Loss is also applied to compute the $loss$ of falling detection task.

% \end{itemize}

\subsection{Training Networks}
CSI-Net was implemented in \textit{Pytorch} 0.4.1 and trained on a Ubuntu server with 4 Titan Xp GPUs. The optimizer is Adam\cite{kingma2014adam} with default parameters ($\beta_1=0.9, \beta_2=0.999$). 
We train the networks for 20 epochs with a minibatch size of 20 and initial learning rate of 0.001. The learning rate is decayed by 10\% at the epoch of $\left \{ 4, 7, 10, 13, 16, 18 \right \}$. 
% In the beginning of each epoch, the training data is shuffled.

% \begin{table}[]
% \centering
% \begin{tabular}{|l|l|l|r|r|r|}
% \hline
% layer name    & output size            & sign recognition          & falling detection         & output size            & layer name    \\ \hline
% avg\_pool1\_1 & $1\times 1 \times 512$ &  $7 \times7$, avg pool &  $7 \times7$, average pool & $1\times 1 \times 512$ & avg\_pool1\_2 \\ \hline
% fc1\_1        & $10$                   & 10-d fc, softmax          & 2-d fc, softmax           & $2$                    & fc1\_2        \\ \hline
% \end{tabular}
% \caption{My caption}
% \label{tab:net2}
% \end{table}
% \subsubsection{Hyper parameters}
% Adam optimizer, learning rate 0.001,  milestone decrease 0.1
% using Adam
% with standard parameters 

%% file: tex/results.tex
\section{Evaluation}\label{sec:evaluation}

In this section, we first present accuracies of CSI-Net on 3 classification tasks, \ie, person recognition, sign recognition and falling detection, and make a comparison with prior work. Then we depict CSI-Net performance on all 4 tasks detailedly. Besides, We show some discoveries about behaviors of CSI-Net processing CSI data. We also list the performance of 9 different backbone networks, including ResNet-152, Inception-V4 and VGG-19. We find that ResNet-18 performed best for our datasets. In addition, we modify the generation stage and propose a new version of CSI-Net.
% The new CSI-Net is lighter but works better, we call it CSI-Net V1.5.

% To our knowledge, this is the first attempt to visualize of DNNs on CSI data. 
% we conduct an attempt %on opening the black-box 
% to figure out the pattern
% that CSI-Net processes CSI with visualizing spatial-encoded maps which generated by transposed convolution layers (generation stage).
% we present the experimental results and explain  the pattern of CSI-Net generating spatial-encoded maps from CSI. To our knowledge, this is the first work that attempts to open the black-box of DNNs processing CSI. 

\subsection{Accuracy Comparison}\label{sec:overall}
We show accuracies of CSI-Net on 3 classification tasks, \ie, person recognition, hand sign recognition and falling detection, in \Table{accuracy}. Meanwhile, accuracies achieved by Support Vector Machine (SVM) and Na\"ive
 Bayes, are listed as the baseline. For SVM, we use LibSVM\cite{CC01a} with radial basis function kernel ($\gamma=1/30$), L2 regularization and penalty of 1. Na\"ive
 Bayes is implemented with assumption that CSI data follows Gaussian distribution. Features used in SVM and Na\"ive
 Bayes are the CSI amplitudes of 30 subcarriers, similar to the CSI-Net. The only difference is that the input size of CSi-Net is $30\time 1\times 1 \times 1$, but the input size of SVM and Na\"ive
 Bayes is $30\times 1$.
%  Due to limited space, we leave more technical details about SVM and Na\"ive Bayes in the Appendix, \Sec{svm} and \Sec{nb}. As aforementioned 
%  %Having emphasized in the data collection section, 
% in \Sec{collection}, we collected CSI under controllable environment and kept CSI parameters unchanged, such as the Tx/Rx settings, orientations and position of subjects. CSI-Net is able to learn the differences in human body or human pose, not the changes in the environment when collecting data. 

% \marginpar{Review 4, for SVM and Naive Bayes, what features were used?}

\begin{table}[h]
\centering
\scriptsize{
\begin{tabular}{llll}
\hline
\multicolumn{1}{c}{\multirow{2}{*}{Method}} & \multicolumn{3}{c}{Accuracy}                                                                          \\ \cline{2-4} 
\multicolumn{1}{c}{}                       & \multicolumn{1}{l|}{Person Recognition} & \multicolumn{1}{l|}{Sign Recognition} & Falling Detection \\ \hline
\multicolumn{1}{l|}{CSI-Net}    & \multicolumn{1}{l|}{\textbf{93.00\%}}              & \multicolumn{1}{l|}{\textbf{100\%}}         & \textbf{96.67\%}          \\ \hline
\multicolumn{1}{l|}{SVM-RBF}               & \multicolumn{1}{l|}{85.28\%}              & \multicolumn{1}{l|}{90.24\%}          & 81.46\%           \\ \hline
\multicolumn{1}{l|}{Na\"ive
 Bayes}           & \multicolumn{1}{l|}{72.97\%}              & \multicolumn{1}{l|}{81.00\%}          & 73.01\%           \\ \hline\hline
\multicolumn{1}{l|}{Prior work}          & \multicolumn{1}{l|} {79.28\%-50\cite{wang2016gait},\
80\%-6\cite{zeng2016wiwho},91\%-11\cite{shi2017smart}
}             & \multicolumn{1}{l|}{$90.2\%$\cite{li2016wifinger}, $93\%$\cite{tan2016wifinger}}          & $90\%$\cite{wang2017wifall}, $91.5\%$ \cite{wang2017rt},$93\%$\cite{palipana2018falldefi}          \\ \hline

\end{tabular}}
\vspace{5pt}
\caption{Classification accuracy comparison.
% ``SVM-RBF'' stands for the C-SVC with radial basis kernel implemented in LibSVM \cite{CC01a}. ``Na\"ive Bayes''  was implemented with assumption that CSI data follows Gaussian distribution. 
 Existing work on person recognition were conducted on different number of subjects, denoted as $ xx\%- \# subjects$. 
% Both ``SVM-RBF" and ``Naive Bayes" used the best hand-craft features \red{[xxxx]} 
}
\label{tab:accuracy}
\end{table}

We see accuracies of CSI-Net on three classification tasks are 93.00\%, 100\% and 96.67\%, respectively. Apparently, CSI-Net outperforms SVM and Na\"ive
 Bayes in all three tasks.
%  The results imply that CSI-Net is able to automatically learn and establish better representation for human sensing than %traditional methods like 
% SVM and Na\"ive Bayes. 
%  The results also verify our assumption that CSI variance caused by static postures can be used for human sensing.
Comparison with closely-related prior work is also shown in \Table{accuracy}, 
% Here we only focus on comparing data processing methods and 
illustrating that CSI-Net is an alternative method to do different tasks. Besides, CSI-Net is able to jointly estimate human biometrics with identification.

\subsection{Biometrics Estimation}\label{sec:bio_result}
To evaluate the results of biometrics estimation, we apply two metrics, \ie, mean average error (\textit{mAE}) and mean square deviation (\textit{mSD}). 
The former is to measure the estimation error of CSI-Net, and the latter for the estimation variance. The \textit{mAE} is computed as follows.
\begin{equation}
e_k = \frac{1}{N_k} \sum_{i=1}^{N_k} \left |  s_{k,i}^*  - s_{k,i}   \right |
\end{equation}
where the $e_k$ represents the average estimation error of the $k$-th subject, $s_{k,i} $ stands for the real $i$-th test value of the $k$-th subject, $s_{k,i}^*$ represents the estimation value, and $N_k$ is the volume of test data of the $k$-th subject. 
The metric, \textit{mAE}, is the \textit{mean} of all average estimation errors ($e_k, k\in[1,30]$). 
\begin{equation}
mAE = \frac{1}{30} \sum_{k=1}^{30} e_k
\end{equation}
\textit{mSD} is computed in a similar way.

\begin{table}[h]
\centering
\begin{tabular}{l|l|l|l|l}
\hline
    & Fat Rate & Muscle Rate & Water Rate & Bone Rate \\ \hline
\textit{mAE} & 1.11     & 1.00        & 0.71        & 0.38      \\ \hline
\textit{mSD} & 0.58     & 0.50         & 0.24       & 0.07      \\ \hline \hline
{min-max range} & [5.0, 30.9]  & [65.2, 89.9] & [49.2, 65.1]& [1.6, 13] \\ \hline

\end{tabular}
\vspace{5pt}
\caption{The mean Average Error (\textit{mAE}) and mean Square Deviation(\textit{mSD}) of four biometrics estimation tasks. 
% Note that comparing to the numerical range of biometrics, CSI-Net produced very small \textit{mAEs} and \textit{mSDs}. To our knowledge, no existing work estimates these biometrics using WiFi signals.
}
\label{tab:bio}
\end{table}

As shown in \Table{bio}, CSI-Net achieves very small \textit{mAEs} and \textit{mSDs} comparing to the numerical range of $30$ subjects' biometrics. Taking the fat rate estimation result as an example, the average estimation error is 1.11, which can be considered as a small error because the min-max range of fat rate is [5.0, 30.9], as listed in \Table{person}. Besides, the estimating variance is 0.58, indicating that CSI-Net has an excellent performance in terms of stability. The results demonstrate that human body does incur strong CSI variance which can be effectively utilized to estimate the human biometrics. 

\begin{figure*}[h]
\includegraphics[width=1\linewidth]{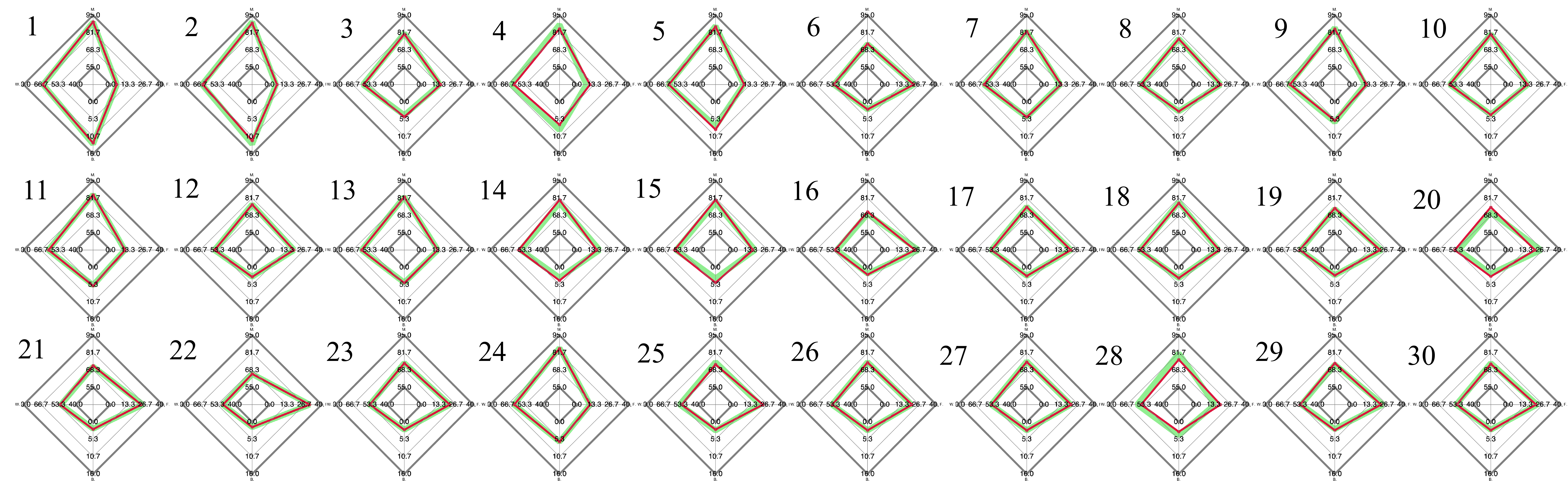}
   \centering
\caption{Radar charts of biometrics estimation for 30 subjects. Four axises (right/up/left/down) correspond to four human biometrics (fat/muscle/water/bone rate), the red lines are estimation, the green lines are ground-truth.}
%The right/up/left/down coordinate corresponds to human fat/muscle/water/bone rate. The orange lines represent the averages of prediction. The blue lines show the ground truth values weighted by a scale\cite{mibodysscale}. We find that most blue lines are overlapped by orange lines, implying that the prediction of CSI-Net is quite close to the ground truth values. 
%We also can see some volunteers, such as the 1st, 2nd and 4th, have low body fat rate and high body muscle rate, so we suggests they enrich their diet. Some volunteers, such as the 6th, the 20th and the 22th, are with high fat rate and low muscle rate, so we suggest they do more exercises. Some other volunteer s(3th,19th and 21th) are with low bone rate, so we suggest they intake more food that with the \textit{calcium}, such as milk, shrimps and beans.}
\label{fig:bio}
% \hspace{-20pt}
\end{figure*}

In \Figure{bio}, we draw a set of radar charts to visualize the estimated biometrics of $30$ subjects. In each chart, the four axises (right/up/left/down) correspond to four human biometrics (fat/muscle/water/bone rate). In the figure, the red lines stand for the average estimation values. The green lines represent the ground truth of biometrics measured by the Mi\textregistered ~ body composition scale\cite{mibodysscale}. The majority of blue lines are perfectly overlapped with the red lines, implying that CSI-Net can estimate biometrics accurately for every subject.
% A few of blue lines appear at the results of the 16th and 28th subjects, implying CSI-Net , .
 
% More importantly, \Figure{bio} shows that CSI can potentially be used to assess body state information for health care.
% For example, the results accurately estimate that some subjects (1st,~2nd,~4th,~5th,~9th,~13th and 24th) are with low fat rate and high muscle rate. 
% %, which indicates they are very thin and should better enrich their diets. 
% On the other hand, some subjects (6th, 16th, 22th and 29th) are with high fat rate and low muscle rate. %, so we suggest they do more exercises for a better body fat rate. 
% Accordingly, we can give them useful suggestions, e.g., enriching the diet or doing more exercises. Using CSI-Net, we can also detect some potential threats to human health. For example, some subjects (3th, 19th and 21th) are with low bone rate, indicating that they should increase the intake of \textit{calcium-rich} foods, such as shrimp, bean and milk. This makes it possible to build daily health maintenance application wherever there is WiFi. 

%%%%%%%%%%%%%%%%%%%%%%%%%%%%%%%%%%%%%%%%%%%%%%%%%%%%%%%%%%%%%%%%%

\subsection{Person Recognition}

\begin{figure*}[h]
\includegraphics[width=0.65\linewidth]{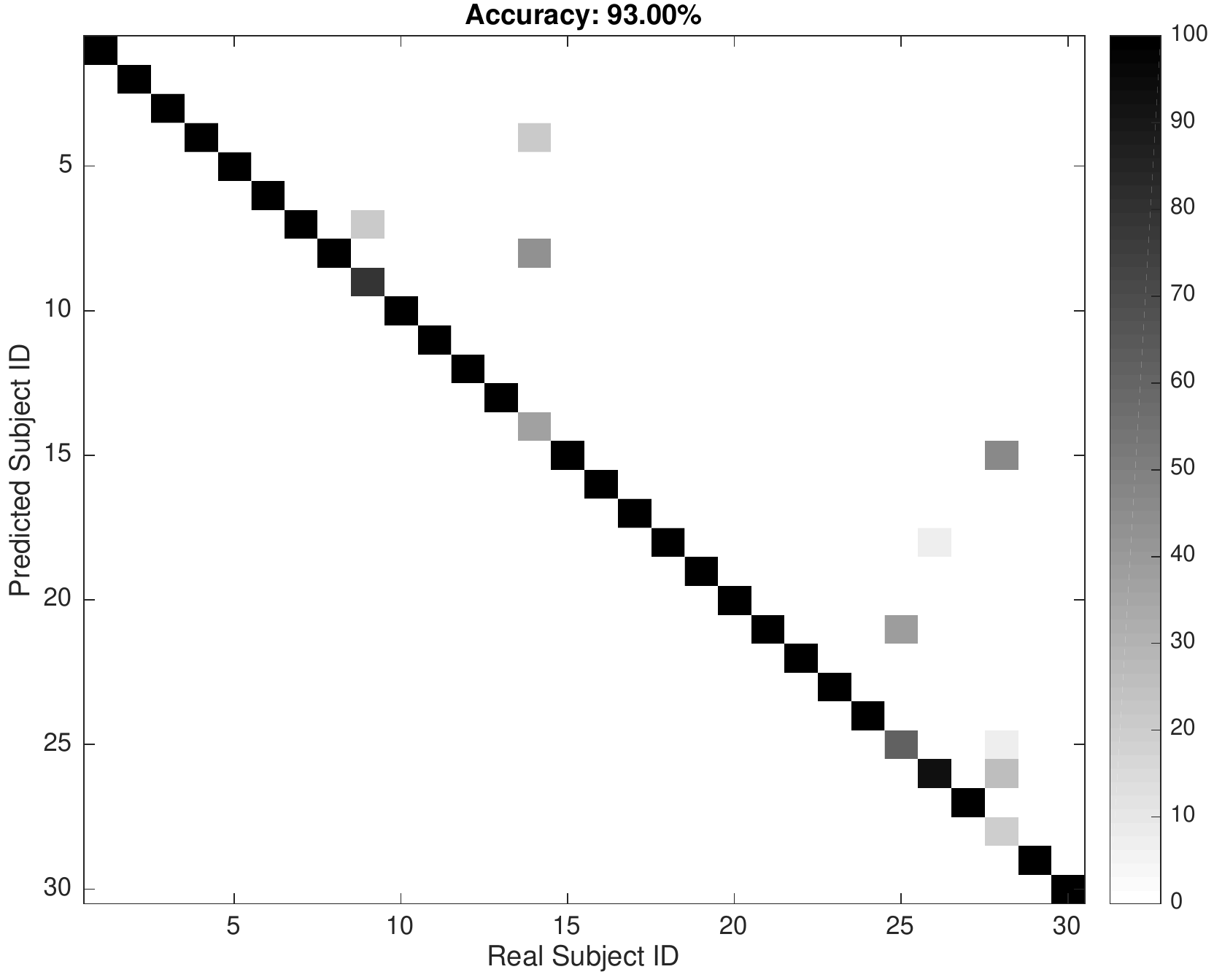}
   \centering
\caption{Confusion matrix of person recognition for $30$ subjects.}
%We use the black to represents percentage of 100\%, and white to represents percentage of 0\%. The gray blocks are represented percentage between 0 and 1, the deeper the gray, the closer it reaches 100\%. From the confusion matrix, we can see CSI-Net can identify most volunteers with accuracy of 100\%. And it preforms unsatisfactorily at the volunteers with ID of $\left\{ 14,20,25,28\right\} $} 
\label{fig:confhuman}
% \hspace{-20pt}
\end{figure*}

We evaluate person recognition results with the confusion matrix in \Figure{confhuman}. The column label is the real human ID and the row label is the predicted human ID. The gray level of block $(i,j)$ represents the ratio that the $j$-th subject is identified as the $i$-th subject by CSI-Net. 
% Thus, the value of each block ranges from 0\% to 100\%. The completely white stands for ratio of 0\% and the completely black stands for ratio of 100\%. 
From the figure, we observe that CSI-Net achieves 100\% accuracy on most of subjects. 
% There are few errors when identifying the subject of $\left\{9,14,25,28\right\} $. Specifically, CSI-Net may incorrectly identify the 8th subject to the 7th subject, 14th subject as the 8th subject, the 25th as the 21th, the 28th as the 15th or 26th. 

\subsection{Sign Recognition}\label{sec:sign_result}

\begin{figure*}[h]
\includegraphics[width=0.65\linewidth]{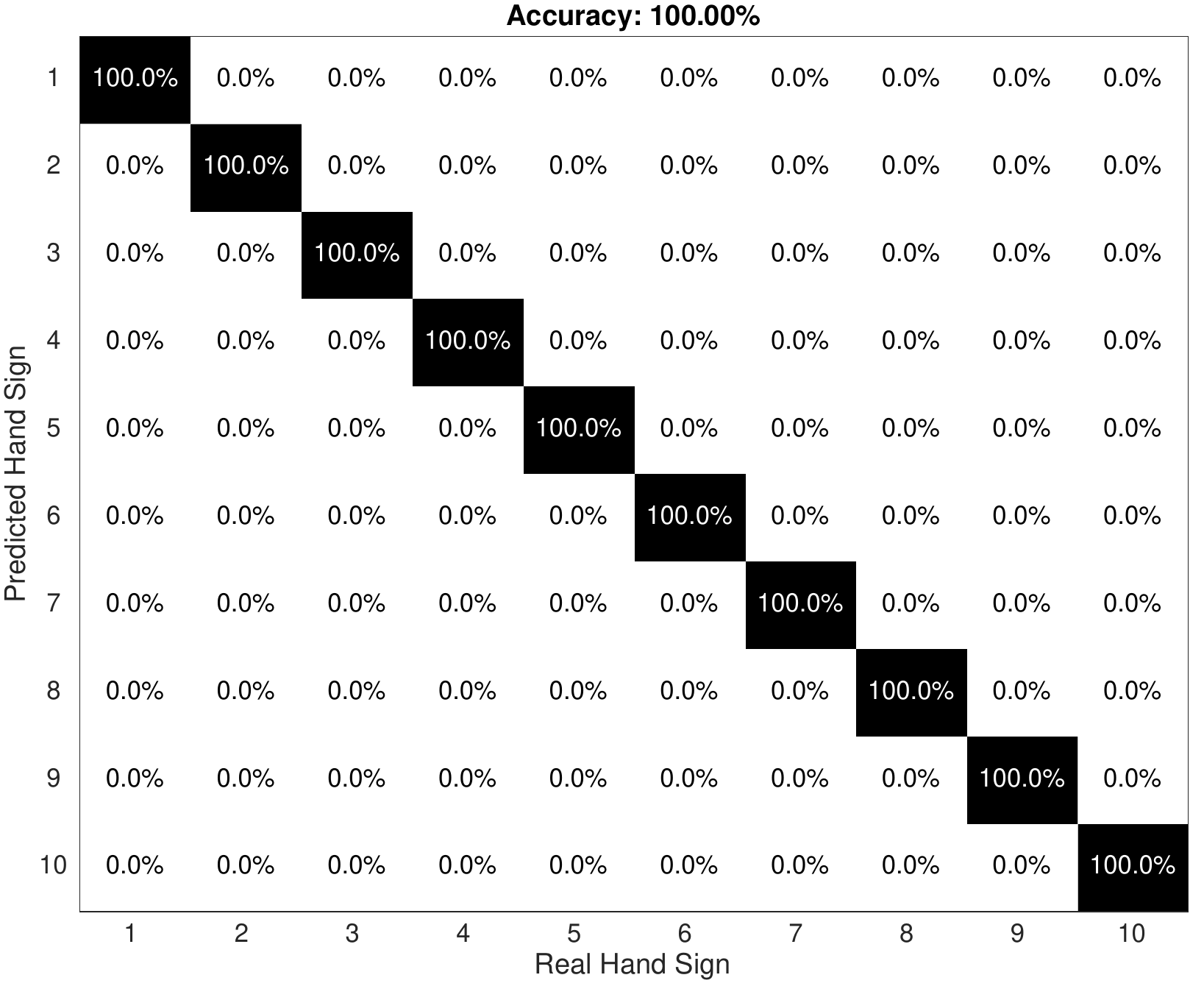}
   \centering
\caption{Confusion matrix of sign recognition. CSI-Net predicted sign of $\left\{0,2,3,4,5,6,7,8,9 \right\}$ with 100\% accuracy. Only 5.6\% error occur where some ``1''s were recognized as ``8''.
}
\label{fig:confsign}
% \hspace{-20pt}
\end{figure*}

We also illustrated the results of hand sign recognition with a confusion matrix in \Figure{confsign}. 
% The column label is the ground truth of hand sign 
The column label is the real hand sign and the row label is the predicted hand sign. In this figure, we find that using our dataset CSI-Net achieves an accuracy of 100\%. This extremely high accuracy demonstrates that micro human postures can result in CSI variances. Meanwhile, even  the variances caused by very similar hand sign gestures can be well recognized by CSI-Net. 
% Because with the same features as applied to CSI-net, the accuracy achieved through SVM is 90.24\%, which is much worse than that of CSI-Net. It is safe to say that CSI-Net is a better classifier to recognize hand sign with CSI data. Besides the power of CSI-Net, we think another reason of achieving high accuracy is that we deployed the transmitter and receiver with a short in-between distance and asked the subject to pose in the line-of-sight, which made CSI very sensitive to the changes caused by gestures. 
%\red{Although the accuracy here is 100\%, we have to say that it is the result achieved for our dataset along with the data augmentation algorithm, depicted in \Algorithm{augmentation}.}

% In the figure, we see that signs of $\left\{0,2,3,4,5,6,7,8,9\right\}$ can be recognized with 100\% accuracy. For `1', the recognition accuracy is 94.4\%. And all the incorrect outputs are `8'. 
% The reason behind may be that the posture of `1' is very similar to that of `8'  as sketched in \Figure{finger} (merely raising up the
% ring finger and little finger from posture of 1 to posture of 8). 
% This extremely high accuracy verifies the ability of CSI-Net recognizing micro and similar human postures. We also think one reason of this high performance is that we deployed transmitter and receiver with a short distance (0.6m), which facilitates the sensitivities of CSI varying with human postures.
% We think the reason of this extremely high performance of CSI-Net recognizing hand signs  
% Even though, the accuracy of recognizing '1' is still high.

%%%%%%%%%%%%%%%%%%%%%%%%%%%%%%%%%%%%%%%%%%%%%%%%%%%%%%%%%%%%%%%%%
\subsection{Falling Detection}\label{sec:falling_result}

The results of falling detection task are listed in \Table{conffall}, where the accuracy of CSI-Net detecting falling is $98.73\%$. In practice, this is a high detection accuracy. Still, CSI-Net wrongly predicts falling activities to standing case with ratio of $1.27\%$, and incorrectly thinks standing case as falling activities  with ratio of $5.41\%$ . This incorrect detection can be solved with a little longer monitoring duration, which allows the system has more samples to vote for a detection results. Here, we do not tend to go deeper in this system optimization, because it is out of main scope of this paper. 

% We then explain the reason why it is more difficult for CSI-Net to detect human falling (2-class problem) than to recognize hand sign (10-class problem). We mainly ascribe it to 3 differences in the data collection setting between these two tasks. First, in falling detection task, the Tx/Rx distance was set to be nearly 3$m$. While in the hand sign detection task, the Tx/Rx distance was set to 0.8$m$, which made CSI more sensitive to human pose change. Second, in the falling detection, we collected CSI from 5 different positions, which was more complex than that of hand sign detection (the data is only collected in one position). Third, in hand sign recognition, subjects were asked to  pose hand in the line-of-sight of Tx/Rx. However, the data for falling detection were mainly collected from the non-line-of-sight paths. It is intuitive that the poses occurring in the line-of-sight may raise more distinguished WiFi variances than that of non-line-of-sight ones.

% The reason behind CSI utilized for 
%All in all, CSI-Net also performs well in human falling detection.  %monitors human falling/standing and produces prediction by the votes of monitoring results. 

\begin{table}[h]
\centering
\begin{tabular}{l|l|l}
\hline
               & Real Falling & Real Standing \\ \hline
Detected Falling  & 1246 (98.73\%)          & 68 (5.41\%)           \\ \hline
Detected Standing & 16 (1.27\%)           & 1189 (94.59\%)          \\ \hline
\end{tabular}
\vspace{5pt}
\caption{Confusion matrix of falling detection.  CSI-Net predicted falling and standing with accuracy of 98.73\% and 94.59\%,  respectively. The error occurs where it predicts 1.27\% of falling to standing, and predicts 5.41\% of standing to falling.}
\label{tab:conffall}
\end{table}
%%%%%%%%%%%%%%%%%%%%%%%%%%%%%%%%%%%%%%%%%%%%%%%%%%%%%%%%%%%%%%%%%%%%

\subsection{Ablation Experiments}\label{sec:ablation}

All above results are obtained with the backbone of ResNet-18~\cite{he2016deep}. As explained in \Sec{net}, CSI-Net is a scalable architecture and can be easily combined with other good CNN networks, such as VGG-net~\cite{simonyan2014very} and Inception~\cite{szegedy2017inception}. Next, we report the results achieved by these backbones and show that ResNet-18 produced the best results for our datasets, shown in \Table{backbone}. In addition, ResNet-18 is the smallest model, shown in \Table{modelsize}. We believe that ResNet-18 gives the best balance between model complexity and data complexity.

% \usepackage{multirow}
\begin{table}[h]
\centering
\begin{tabular}{c|c|c|c|c|c|c}
\hline
\multirow{2}{*}{backbone} & \multicolumn{2}{c|}{Person Recognition} & \multicolumn{2}{c|}{Sign Recognition} & \multicolumn{2}{c}{Falling Detection} \\ \cline{2-7} 
                          & acc~(train)         & acc~(test)         & acc~(train)         & acc~(test)         & acc~(train)         & acc~(test)          \\ \hline
\textbf{ResNet-18}                 & 100\%          & \textbf{93.00\%}          & 100\%         & \textbf{100\%}          & 98.13\%          & \textbf{96.67\%}         \\ \hline
ResNet-34                 & 100\%          & 85.10\%          & 98.74\%         & 72.69\%          & 96.02\%         & 94.40\%          \\ \hline
\textit{ResNet-50}                 & 99.99\%          & 87.51\%          & 99.99\%          & \textbf{100\%}          & 95.77\%          & 93.85\%         \\ \hline
\textit{ResNet-101}                 & 100\%          & 88.41\%          & 99.96\%          & \textbf{99.03\%}          & 96.14\%          & 93.33\%          \\ \hline
ResNet-152                 &  100\%         & 88.61\%          & 100\%         & \textbf{100\%}          & 94.07\%          & 89.68\%          \\ \hline
\textit{Inception-V3}                 & 99.94\%          & 82.63\%          & 100\%          & \textbf{100\%}          & 97.04\%          & \textbf{96.35\%}          \\ \hline
\textit{Inception-V4}                 &  100\%          & 87.60\%          & 100\%          &  \textbf{100\%}          & 96.11\%          & 93.01\%         \\ \hline
VGG-16                 & 30.03\%         & 18.70\%          & 80.31\%          & 46.67\%          & 87.63\%          & 83.56\%          \\ \hline
VGG-19                 &  19.03\%         & 16.00\%          & 75.27\%          & 42.30\%          & 88.83          & 85.15\%          \\ \hline
\end{tabular}
\vspace{5pt}
\caption{Result of different backbones. ``acc(train)'' and ``acc(test)'' represent training accuracy and test accuracy, respectively.
}
\label{tab:backbone}
\end{table}

\begin{table}[h]
\centering
\begin{tabular}{c|c|c|c}
\hline
 backbone& Person Recognition & Sign Recognition &  Falling Detection\\ \hline
 \textbf{ResNet-18}& \textbf{88MB} & \textbf{61MB} & \textbf{61MB} \\ \hline
 \textit{ResNet-34}& 127MB & 102MB & 102MB \\ \hline
 ResNet-50& 189MB & 176MB &176MB  \\ \hline 
 ResNet-101& 798MB & 366MB & 366MB  \\ \hline
 ResNet-152& 878MB & 423MB &  423MB\\ \hline
 \textit{Inception-V3}& 90MB & 90MB  & 90MB \\ \hline
 Inception-V4& 164MB & 164MB & 164MB \\ \hline
 VGG-16& 975MB & 531MB & 531MB \\ \hline
 VGG-19& 995MB & 552MB & 552MB \\ \hline 
\end{tabular}
\vspace{5pt}
\caption{Saved model size.
}\label{tab:modelsize}
\end{table}

%%%%%%%%%%%%%%%%%%%%%%%%%%%%%%%%%%%%%%%%%%%%%%%%%%%%%%%%%%%%%%%%%%%%
% \subsection{Discovery}\label{sec:discovery}

\subsection{Deconvolution or Interpolation} \label{sec:v15}
In the generation stage of CSI-Net, 8 stacked transposed convolution layers convert the CSI data to a tensor with height/width $224\times 224$, which is approximating to conventional input size of CNNs. 
% This idea of applying stacked transposed convolution layers is derived from DC-GAN~\cite{radford2015unsupervised}, a work making 1D tensor to an vivid image. 
However, considering the computation cost of these layers, we explored simpler but still efficient ways to reshape CSI into image-like tensors for CNNs. Then, we had a better deep architecture for our tasks. Next, we shown this update.

The first method is to resize CSI tensor from size of $30\times 1\times 1$ to size of $30 \times 224 \times 224$ with \textit{bilinear} interpolation, whose performance is shown in \Table{v15}. Surprisingly, this simple method achieves high accuracy in all tasks, which inspired us to reconsider the function of interpolation operation.
% We think the strength of interpolation is to reduce the amount of parameters to be optimized, which would reduce the risk of over-fitting. However, the weakness of interpolation is that it cannot introduce any nonlinear mapping ability into CSI-Net. So, we try an intuitive idea: combining the transposed convolution layer and interpolation operation to generate spatially-encoded image-like tensors. 
% Recalling that we have 8 transposed convolution layers in CSI-Net, 
We also replace the 2nd, 4th, 6th, and 8th layers with bilinear interpolations, resulting in a lighter CSI-Net. It outperforms the original CSI-Net with a notable increase in the accuracy. The results are shown in \Table{v15}. 
% We call the new architecture CSI-Net V1.5.

\begin{table}[h]
\small{
\begin{tabular}{c|c|c|c|c|c|c}
\hline
\multirow{2}{*}{Method} & \multicolumn{2}{c|}{Person Recognition} & \multicolumn{2}{c|}{Sign Recognition} & \multicolumn{2}{c}{Falling Detection} \\ \cline{2-7} 
                          & acc~(train)         & acc~(test)         & acc~(train)         & acc~(test)         & acc~(train)         & acc~(test)          \\ \hline
Transposed Convolution (TC)                & 100\%          & 93.00\%          & 100\%         & 100\%          & 98.13\%          & 96.67\%          \\ \hline
Interpolation (I)                 & 100\%          & 91.36\%          & 100\%         & 100\%          & 98.12\%          & 93.45\%          \\ \hline
CSI-Net V1.5 (TC+I)               & 100\%          & 94.42\%          & 100\%          & 100\%          & 98.07\%          & 97.73\%          \\ \hline
\end{tabular}}
\vspace{5pt}
\caption{Three methods to make CSI sample image-like tensor. Interpolation means directly resizing $30\times 1\times 1$ CSI tensor to be $30\times 224\times 224$ with bilinear interpolation. 
% Transposed Convolution represents original CSI-Net. 
% CSI-Net V1.5 (TC+I) is to replace the 2nd, 4th, 6th and 8th transposed convolution layers with interpolation. It is the best of all.
}
\label{tab:v15}
\end{table}

%% file: tex/appendix.tex
\clearpage
\section*{Appendix}\label{sec:appendix}

\subsection*{Detailed CSI-Net Parameters}\label{sec:layers}

\begin{figure*}[h]
\includegraphics[width=1\linewidth]{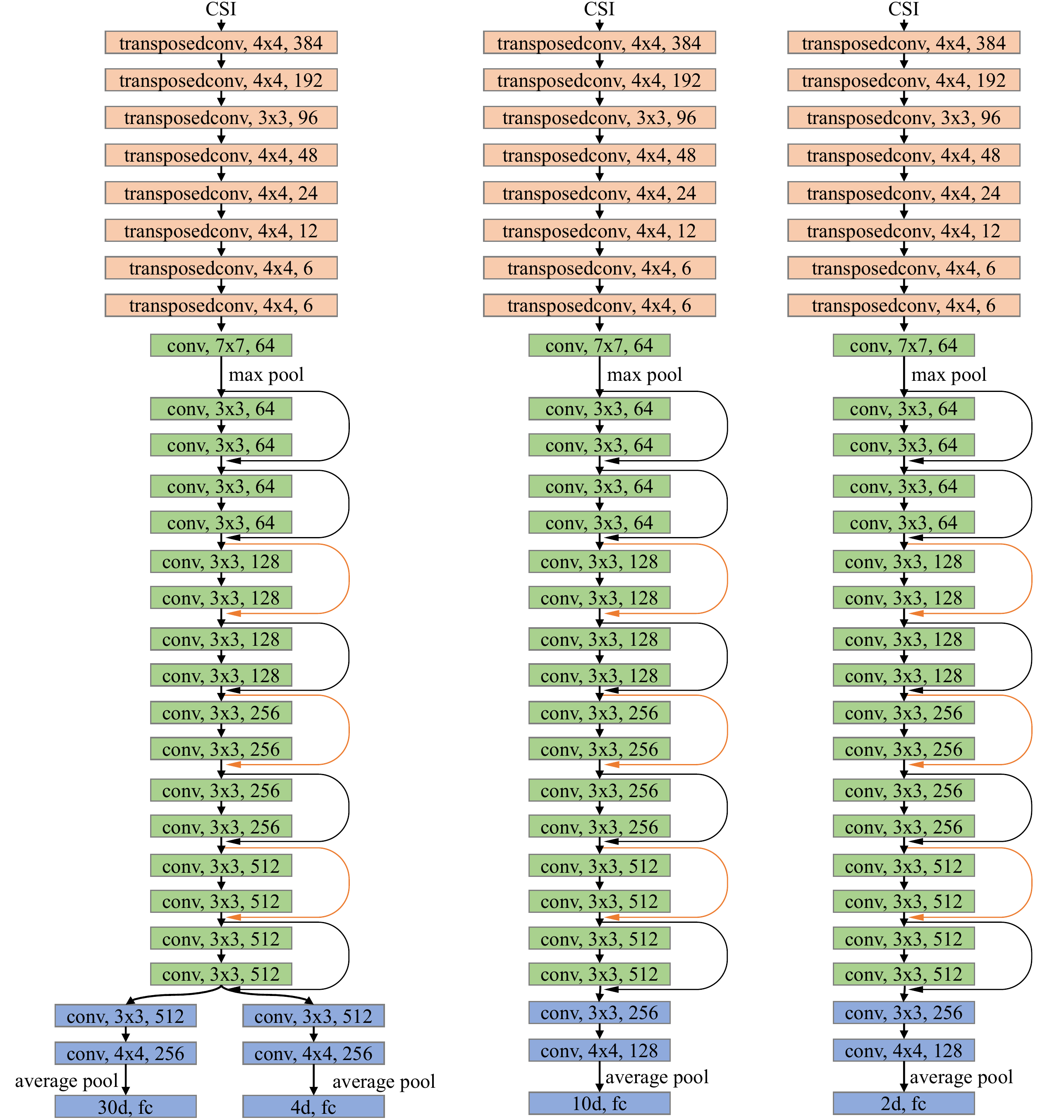}
   \centering
\caption{Sketches of CSI-Net architectures for tasks of, \textbf{Left}: joint biometrics estimation and person recognition. \textbf{Middle}: hand sign recognition, and \textbf{Right:} falling detection. The orange shortcuts increase channels. Please see \url{https://github.com/geekfeiw/CSI-Net} for a Pytorch implementation.}
\label{fig:net_flow}
% \hspace{-20pt}
\end{figure*}

\clearpage
\subsection*{Subjects' Information}
\input{tex/human_table}

%% file: tex/human_table.tex
\begin{table}[h]
\centering
\scriptsize{
\begin{tabular}{llcccccc}
\hline
No. & Sex    & \multicolumn{1}{l}{Fat Rate (\%)} & \multicolumn{1}{l}{Muscle Rate (\%)} & \multicolumn{1}{l}{Water Rate (\%)} & \multicolumn{1}{l}{Bone Rate (\%)} & \multicolumn{1}{l}{Height (inches)} & \multicolumn{1}{l}{Weight (lbs)} \\ \hline
1   & Male   & 5.0                               & 89.7                                 & 65.1                                & 13.0                               & 70.5                                & 113.8                             \\
2   & Male   & 5.0                               & 89.9                                 & 65.1                                & 13.0                               & 66.1                                & 107.9                             \\
3   & Male   & 14.3                              & 81.2                                 & 58.7                                & 4.1                                & 68.1                                & 141.2                             \\
4   & Male   & 7.3                               & 87.8                                 & 63.5                                & 8.7                                & 68.5                                & 121.0                             \\
5   & Male   & 8.4                               & 86.7                                 & 62.8                                & 7.5                                & 66.9                                & 118.9                             \\
6   & Female & 22.9                              & 72.8                                 & 52.8                                & 2.3                                & 62.6                                & 104.7                             \\
7   & Male   & 12.7                              & 82.7                                 & 59.8                                & 4.7                                & 70.9                                & 143.7                             \\
8   & Male   & 18.9                              & 76.9                                 & 55.6                                & 2.9                                & 71.3                                & 161.7                             \\
9   & Male   & 9.9                               & 85.3                                 & 61.7                                & 6.2                                & 71.3                                & 134.2                             \\
10  & Male   & 15.0                              & 80.5                                 & 58.2                                & 3.9                                & 65.0                                & 127.9                             \\
11  & Male   & 11.0                              & 84.3                                 & 61.0                                  & 5.5                                & 70.5                                & 138.3                             \\
12  & Male   & 18.9                              & 76.8                                 & 55.6                                & 2.9                                & 66.9                                & 148.2                             \\
13  & Female & 12.7                              & 82.6                                 & 59.8                                & 4.7                                & 62.6                                & 114.5                             \\
14  & Female & 17.5                              & 78.0                                 & 56.5                                & 3.2                                & 63.8                                & 105.4                             \\
15  & Female & 16.9                              & 78.2                                 & 56.9                                & 3.4                                & 66.5                                & 109.1                             \\
16  & Female & 26.2                              & 69.4                                 & 50.6                                & 1.9                                & 63.4                                & 121.9                             \\
17  & Male   & 20.9                              & 75.0                                 & 54.2                                & 2.6                                & 68.5                                & 161.2                             \\
18  & Female & 17.2                              & 78.3                                 & 56.7                                & 3.3                                & 64.2                                & 105.9                             \\
19  & Female & 22.5                              & 73.2                                 & 53.1                                & 2.4                                & 63.0                                & 104.8                             \\
20  & Male   & 25.1                              & 70.9                                 & 51.3                                & 2.0                                & 71.7                                & 190.5                             \\
21  & Female & 24.9                              & 70.9                                 & 51.4                                & 2.1                                & 61.8                                & 106.0                             \\
22  & Female & 30.9                              & 65.2                                 & 49.2                                & 1.6                                & 60.6                                & 118.4                             \\
23  & Female & 22.0                              & 73.7                                 & 53.5                                & 2.4                                & 62.2                                & 102.1                             \\
24  & Male   & 10.1                              & 85.2                                 & 61.6                                & 6.1                                & 69.3                                & 128.9                             \\
25  & Female & 20.9                              & 74.4                                 & 54.2                                & 2.6                                & 62.6                                & 110.6                             \\
26  & Female & 20.8                              & 74.4                                 & 54.2                                & 2.6                                & 65.4                                & 110.6                             \\
27  & Male   & 20.8                              & 75.0                                 & 54.2                                & 2.6                                & 68.9                                & 162.5                             \\
28  & Male   & 14.4                              & 81.1                                 & 58.6                                & 4.1                                & 66.9                                & 135.4                             \\
29  & Male   & 23.5                              & 72.5                                 & 52.4                                & 2.2                                & 72.8                                & 187.9                             \\
30  & Male   & 22.3                              & 73.6                                 & 53.2                                & 2.4                                & 65.0                                & 152.7        \\ \hline                   
\end{tabular}}
\vspace{5pt}
\caption{The information of 30 recruited subjects. All biometrics were measured using a body composition scale which embedded the module of Bioelectrical Impedance Analysis(BIA)\cite{kyle2004bioelectrical}. Subjects were asked to stand on the scale after removing shoes and socks. All metrics were recorded when the reading of scale is stable.}
\label{tab:person}

\end{table}